\documentclass[10pt, a4paper]{article}
\usepackage{lrec-coling2024} % this is the new style

\usepackage{booktabs}
\usepackage{algorithm}
\usepackage{algorithmic}
\usepackage{enumitem}
\usepackage{amsmath,bm}
\usepackage{pifont}
\usepackage{array}
\usepackage{multirow}
\usepackage{graphicx}
\usepackage{subfig}
\usepackage{amssymb}
\usepackage{color, soul}
\usepackage{color}
\usepackage{colortbl}
\usepackage{tabularx}
\usepackage{tcolorbox}

\usepackage{pgfplots}
\usepackage{scalefnt}
\definecolor{color1}{RGB}{255,128,0}
\definecolor{color2}{RGB}{0,0,255}
\definecolor{color3}{RGB}{227,207,87}
\definecolor{color4}{RGB}{199,97,20}
\definecolor{color5}{RGB}{34,139,34}

\usepackage{enumitem}
\setenumerate[1]{itemsep=0pt,partopsep=0pt,parsep=\parskip,topsep=5pt}
\setitemize[1]{itemsep=0pt,partopsep=0pt,parsep=\parskip,topsep=5pt}
\setdescription{itemsep=0pt,partopsep=0pt,parsep=\parskip,topsep=5pt}

\title{TopicDiff: A Topic-enriched Diffusion Approach for Multimodal Conversational Emotion Detection}

\name{Jiamin Luo$^*$, Jingjing Wang$^*$\thanks{\@ \@  $^*$Equal Contribution.}, Guodong Zhou$^\dagger$\thanks{\@ \@  $^\dagger$Corresponding Author: Guodong Zhou.}} 

\address{School of Computer Science and Technology, Soochow University, China \\
         No.1, Shizi Street, Suzhou City, Jiangsu Province, China \\
         20204027003@stu.suda.edu.cn, \{djingwang, gdzhou, \}@suda.edu.cn\\}

\abstract{
Multimodal Conversational Emotion (MCE) detection, generally spanning across the acoustic, vision and language modalities, has attracted increasing interest in the multimedia community. Previous studies predominantly focus on learning contextual information in conversations with only a few considering the topic information in single language modality, while always neglecting the acoustic and vision topic information. On this basis, we propose a model-agnostic \textbf{Topic}-enriched \textbf{Diff}usion (TopicDiff) approach for capturing multimodal topic information in MCE tasks. Particularly, we integrate the diffusion model into neural topic model to alleviate the diversity deficiency problem of neural topic model in capturing topic information. Detailed evaluations demonstrate the significant improvements of TopicDiff over the state-of-the-art MCE baselines, justifying the importance of multimodal topic information to MCE and the effectiveness of TopicDiff in capturing such information. Furthermore, we observe an interesting finding that the topic information in acoustic and vision is more discriminative and robust compared to the language.
 \\ \newline \Keywords{Multimodal Conversational Emotion Detection, Multimodal Topic, Diffusion Model} }

\begin{document}

\maketitleabstract

\section{Introduction}
\label{sec:intro}
\begin{figure}
\begin{center}
    \subfloat{
 \includegraphics[scale=0.43]{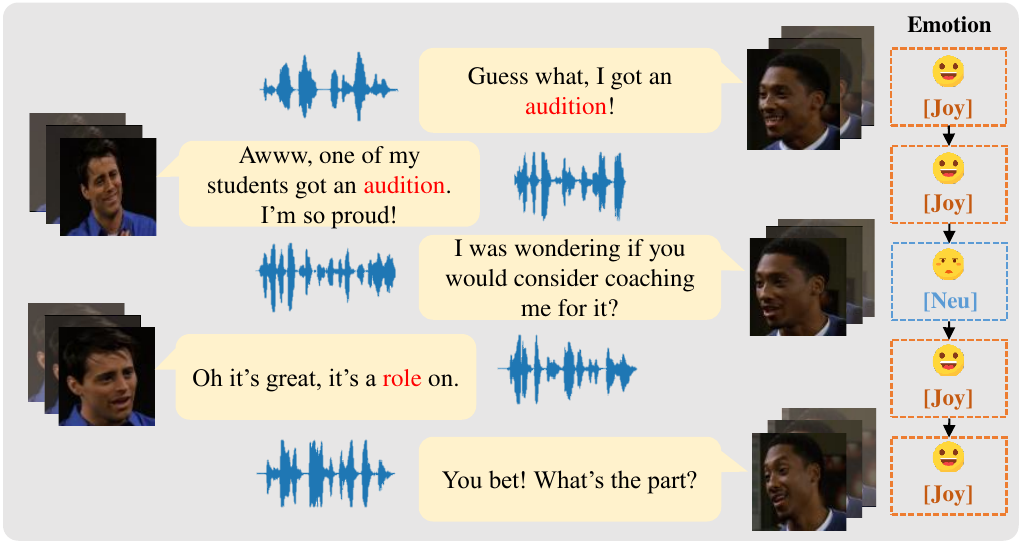}}
\caption{A multimodal conversational example from MELD dataset to illustrate the importance of multimodal topic information, where each utterance contains acoustic spectrum, video frame, language and corresponding emotion label.}
\label{fig:example}
\end{center}
\end{figure}

Multimodal Conversational Emotion (MCE) detection constitutes a crucial task in the fields of natural language processing (NLP)~\cite{PoriaCHMZM17,MajumderPHMGC19,GhosalMPCG19} and multimodal processing~\cite{HazarikaPMCZ18,HuLZJ20}, with the objective of automatically detecting emotions in each utterance within a conversation. As exemplified in Figure \ref{fig:example}, an utterance (including acoustic spectrum, video frame and language ``\emph{Guess what, I got an audition}'') generated by a speaker could be labeled with emotion \emph{joy}. Over the past decade, MCE has a significant impact on the development of empathetic systems and various potential applications, including opinion mining, healthcare and intelligent assistants~\cite{JiaoYKL19,CaoTIPAS19}. Existing studies have primarily focused on employing RNN-variants (e.g., LSTM or GRU)~\cite{PoriaCHMZM17,MajumderPHMGC19}, graph-based~\cite{GhosalMPCG19} and transformer-based~\cite{ShenCQX21} models to capture the contextual information of individual utterances during a conversation.

Recently, a few studies realize the importance of topic information in conversation tasks~\cite{WangWSLLSZZ20,ZhuP0ZH20}. They assert that the utilization of topic information in conversational contexts contributes to fully mining the global clues of utterances during the whole conversation. For instance, conversations regarding the \emph{funeral} topic tend to elicit \emph{sadness}, while those about the \emph{wedding} topic tend to elicit \emph{happiness}. Therefore, considering the topic information in a conversation could improve our comprehension of content, context and intent within the conversation. However, existing studies on incorporating the topic information into conversation tasks solely rely on single language modality, without taking the topic information present in both acoustic and vision into account.  

In this paper, we argue that the multimodal topic information is crucial for boosting the performance of emotion detection in multimodal conversation. By analyzing various modal topic information, such as tones, facial expressions, and body postures, we can better detect the emotions. As illustrated in Figure \ref{fig:example}, we observe a multimodal conversation between two speakers, where the language clues ``\emph{getting an audition}'' and ``\emph{role}'', indicating the topic related to \emph{good job opportunities} that could potentially aid in detecting the emotions of speakers. Moreover, the acoustic spectrums with \emph{excited} tones and video frames with \emph{smile} facial expression generated by two speakers indicate a positive (i.e., \emph{joy}) emotion throughout the conversation, which are also useful to discriminate the emotions. Besides, as previously mentioned conversations related to \emph{funeral} and \emph{wedding} topics, the tones (frequency and intensity variations in tones) and facial expressions (smiling or crying face in video frames) can precisely detect the emotions in the whole conversations. Therefore, we believe that a well-behaved approach should consider topic information within multiple modalities in multimodal conversation to enable the precise detection of emotions.  

However, traditional neural topic models~\cite{WangHZHXYX20,JinZLDB21} generally adopt the architecture of variational auto-encoder (VAE)~\cite{KingmaW13} to capture the topic information inside language, which may suffer from the problem of semantic diversity deficiency~\cite{GaoBLLZS19}. More seriously, acoustic and vision semantics are often more sparse compared to language~\cite{HeCXLDG22}. This may further exacerbate the problem of diversity deficiency in topic information. With this in mind, traditional neural topic models are potentially not well-suited for capturing multimodal topic information. Recently, the diffusion model~\cite{SongE19} has attracted increasing attention due to their ability to sufficiently capture the diverse information from multimodal semantic spaces, achieving remarkable performance in various multimodal tasks~\cite{ruan2022mmdiffusion,huang2023collaborative}. Therefore, we believe that a better-behaved topic mining approach should consider integrating the diffusion model to capture the multimodal topic information.

To tackle the aforementioned challenges, this paper proposes a new model-agnostic \textbf{Topic}-enriched \textbf{Diff}usion (TopicDiff) approach to capture the multimodal topic information for addressing MCE. Specifically, we first integrate the diffusion model into neural topic model to alleviate the problem of semantic diversity deficiency. Then, we leverage three TopicDiff modules to capture the topic information inside acoustic, vision, and language modalities, respectively. Finally, we jointly train the basic MCE approaches alongside three TopicDiff modules under the architecture of multi-task learning. Detailed evaluations demonstrate that our TopicDiff approach achieves significant performance improvements compared to the state-of-the-art MCE baselines. Overall, the main contributions of this paper are summarized as follows: 
\begin{itemize}
\setlength{\itemsep}{0pt}
\setlength{\parsep}{0pt}
\setlength{\parskip}{0pt}
    \item We are the first to consider the multimodal topic information for boosting the performance of conversational emotion detection.
    \item We propose a TopicDiff approach to capture the multimodal topic information in MCE, which first attempts to integrate the diffusion model into neural topic models for alleviating the diversity deficiency problem of them in capturing the topic information. Since TopicDiff is model-agnostic, it can be easily extended to MCE approaches.
    \item We conduct detailed experiments on our topic-density M3ED$^*$ and two public topic-sparsity MELD, IEMOCAP datasets, and the results on these datasets demonstrate that TopicDiff achieves significant improvements compared to state-of-the-art MCE baselines. This fully justifies the importance of multimodal topic information and the effectiveness of our TopicDiff approach in capturing such information.
    \item We observe an interesting finding that the topic information in acoustic and vision is more discriminative and robust than that in language.  
\end{itemize}

\section{Related Work}
\textbf{Multimodal Conversational Emotion Detection.} Previous studies in MCE focus on capturing contextual information due to the unique structure of conversations, which can be broadly categorized into RNN-, graph-, and knowledge-based approaches. RNN-based approaches leverage RNN variant networks to model contextual information~\cite{PoriaCHMZM17,HazarikaPZCMZ18} or track emotional states~\cite{HazarikaPMCZ18,MajumderPHMGC19} in conversations. Graph-based approaches leverage graph networks to model the interaction of speakers~\cite{GhosalMPCG19,IshiwatariYMG20}, model multimodal dependencies and speaker dependencies~\cite{HuLZJ20,JoshiBJSM22,LianCSLT23}, and understand the conversational contexts~\cite{HuWH20,HuHWJM22}. Knowledge-based approaches leverage external knowledge to guide contextual modeling, such as commonsense knowledge~\cite{GhosalMGMP20}, psychological knowledge~\cite{Li00W21}, knowledge from pre-trained language models~\cite{ShenCQX21,ZhuP0ZH20}, while these approaches only consider language modality. 

In summary, all the above studies ignore the multimodal topic information, which can be leveraged to boost the performance of emotion detection in multimodal conversation.

\begin{figure*}
\begin{center}
    \subfloat{
 \includegraphics[scale=0.56]{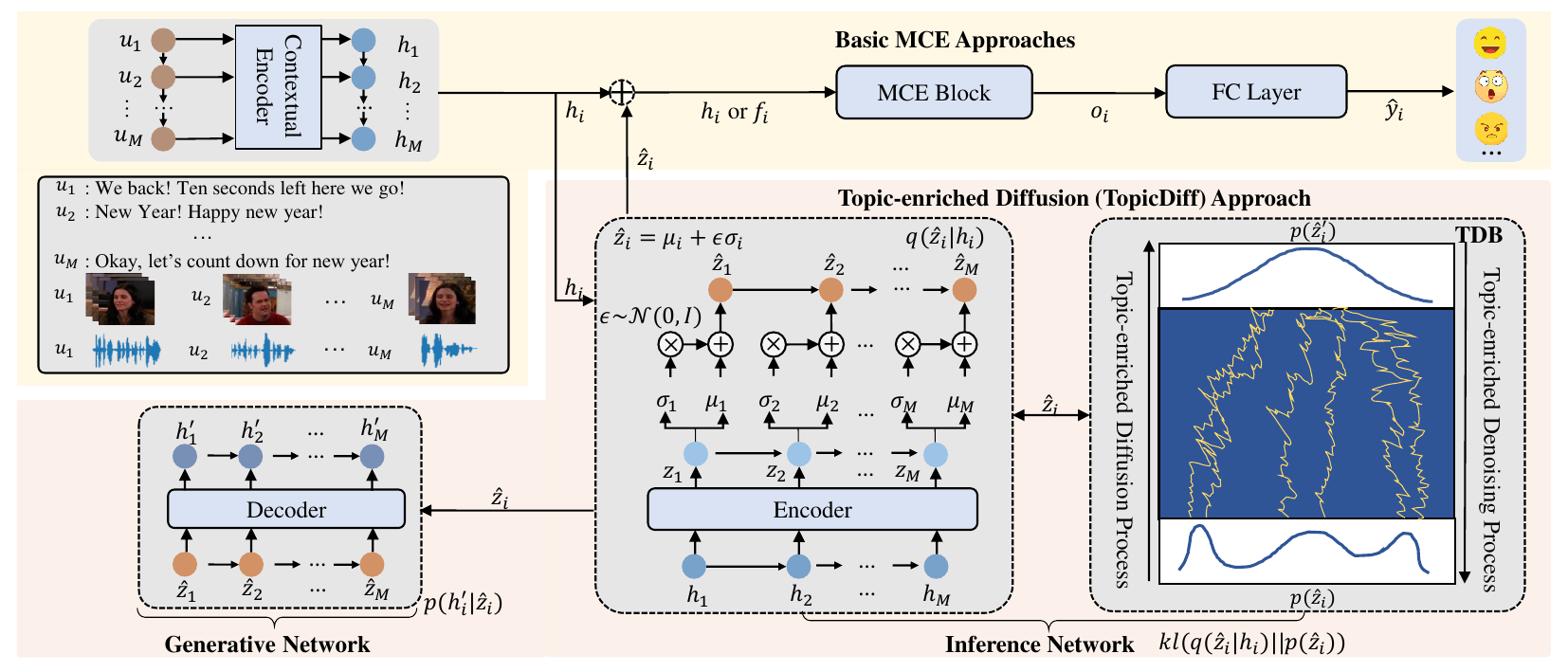}}
\caption{The overall architecture of our model-agnostic Topic-enriched Diffusion (TopicDiff) approach for MCE, where TDB represents Topic-enriched Diffusion Block consisting of Topic-enriched Diffusion Process and Topic-enriched Denoising Process.}
\label{fig:model}
\end{center}
\end{figure*}

\textbf{Neural Topic Model.} Based on the architecture of variational auto-encoder (VAE)~\cite{KingmaW13}, \citet{MiaoYB16} propose neural topic models to capture the topic information within languages. Then, some improved studies attempt to employ various strategies to address problems in language topic modeling, such as adversarial training~\cite{WangHZHXYX20}, pre-trained semantic embeddings~\cite{JinZLDB21}, reinforcement learning~\cite{GuiLPZXH19} and multi-task learning~\cite{WangWSLLSZZ20}. Besides, a few studies~\cite{abs-2010-13373,AnWLZ20} consider incorporating the multimodal topic information, while these efforts are not designed to conversation scenarios and hence are not appropriate for our MCE tasks.

Different from all above studies, we focus on multimodal conversation scenarios and design an TopicDiff approach to mine multimodal topic information, which could mitigate the diversity deficiency problem of traditional neural topic models in capturing topic information.

\textbf{Diffusion Model.} Recently, diffusion models are the state-of-the-art latent variable models in generative modeling~\cite{abs-2209-00796}, which have two main paradigms, i.e., Denoising Diffusion Probabilistic Models (DDPM)~\cite{HoJA20} and Score-based Generative Models (SGM)~\cite{SongE19}. At present, there are many studies on diffusion models~\cite{VahdatKK21,BaranchukVRKB22,RombachBLEO22}, of which \citet{VahdatKK21} train SGMs with VAE to explore the effects of the diffusion model on image generation. \citet{ruan2022mmdiffusion} and \citet{huang2023collaborative} leverage the diffusion model to address the multimodal tasks and achieve promising results.

Overall, the diffusion model has shown a remarkable ability in capturing diverse information from latent spaces. Inspired by this, we first attempt to take advantage of the diffusion model and integrate it into neural topic model to mitigate the diversity deficiency problem, thereby sufficiently capturing the multimodal topic information in MCE.

\section{Approach} 
In this section, we formulate the MCE task as follows. A conversation can be defined as a sequence of utterances $\bm{u}=\left \{ \bm{u}_1, \bm{u}_2, ..., \bm{u}_M \right \}$, where $M$ is the number of utterances. Each utterance involves three sources of utterance-aligned data corresponding to acoustic ($\boldsymbol{a}$), vision ($\boldsymbol{v}$) and language ($\boldsymbol{l}$), which can be formulated as: $\bm{u}_i=\left \{\bm{u}_i^a, \bm{u}_i^v, \bm{u}_i^l \right \}$, where $\bm{u}_i^a$, $\bm{u}_i^v$ and $\bm{u}_i^l$ denote the raw feature representation of $\bm{u}_i$ from acoustic, vision and language, respectively. The MCE task aims to predict the emotional status label for each utterance $\bm{u}_i$ in the conversation based on the available information from all three modalities. 

In this paper, we propose a model-agnostic \textbf{Topic}-enriched \textbf{Diff}usion (TopicDiff) approach for MCE. Figure \ref{fig:model} shows the overall architecture of TopicDiff, consisting of an inference network with a topic-enriched diffusion block (TDB), and a generative network. Before introducing TopicDiff, we first give an overview of the basic MCE approaches.

\subsection{Basic MCE Approaches} 
The basic MCE approaches mainly focus on leveraging RNN-variants (e.g., GRU), graph-based or transformer-based models to model contextual information of each conversation, and the general workflow of them is described as follows. Firstly, each utterance $\bm{u}_i$ is encoded by different modal contextual encoders to obtain context-aware representation $\bm{h}_i$. Various MCE approaches employ different contextual encoders for each modality, which we denote as $\rm CTEncoder$ for uniformity. Thus, the context-aware representation $\bm{h}_i$ of each utterance $\bm{u}_i$ is formulated as follows:
\begin{equation}
\label{eq:hi}
    \bm{h}_i = {\rm CTEncoder}(\bm{u}_i) 
\end{equation}
where $\bm{h}_i=\left \{\bm{h}_i^a, \bm{h}_i^v, \bm{h}_i^l \right \}$ is serving as the uniform representation of acoustic, vision, and language

Subsequently, $\bm{h}_i$ is fed into various MCE blocks (e.g., MMGCN employs a spectral domain graph convolutional network to encode the multimodal contextual representations), where $\bm{h}_i$ is transformed into an intermediate representation $\bm{o}_i$. For clarity, we use ``MCE Block'' to represent the core and complex components of various MCE approaches as shown in Figure \ref{fig:model}.

Finally, the emotion prediction $\hat{y}_i$ is made by passing $\bm{o}_i$ through a fully connected (FC) layer. The basic MCE approaches are trained via the cross-entropy loss function, which can be defined as:  
\begin{equation}
    \mathcal{L}_{mce} = - \frac{1}{\sum_{n=1}^Nc(n)}\sum_{j=1}^N\sum_{i=1}^{c(n)}y_{j,i}^n\log \hat{y}_{j,i}^{n}
\end{equation}
where $N$ is the total number of conversations in the training set. $c(n)$ is the number of utterances in conversation $n$. $y_{j,i}^n$ and $\hat{y}_{j,i}^{n}$ are the expected class label and the predicted emotion label of utterance $i$ in conversation $j$, respectively.

\subsection{Topic-enriched Diffusion Approach}
To capture the topic information inside acoustic, vision, and language, we propose a Topic-enriched Diffusion (TopicDiff) approach. TopicDiff is essentially a variant of neural topic model~\cite{MiaoYB16}, and thus also consists of two main components, i.e., inference network and generative network as illustrated in Figure \ref{fig:model}. Particularly, we design a \textbf{Topic-enriched Diffusion Block (TDB)}, which is then integrated into the inference network for alleviating the diversity deficiency problem of traditional vae-based neural topic model. Specifically, we leverage three TopicDiff modules to capture the topic information inside acoustic, vision, and language on the basis of basic MCE approaches, respectively.  

\textbf{Inference Network} is leveraged to map the context-aware representation $\bm{h}_i$ in Eq.(\ref{eq:hi}) to a low-dimension latent topic representation $\bm{\hat{z}}_i$.\footnote{After obtaining the context-aware representation $\bm{h}_i$, we incorporate latent topic representation $\bm{\hat{z}}_i$ via TopicDiff, without modifying the overall architecture of MCE approaches. Therefore, we say that our TopicDiff approach is model-agnostic.} Specifically, an MLP encoder layer is first employed to extract features from $\bm{h}_i$ to obtain the output $\bm{z}_i$. And then $\bm{z}_i$ can be fed into two different fully connected layers $f_{\mu}$ and $f_{\sigma}$ to estimate the mean $\bm{\mu}_i$ and standard deviation $\bm{\sigma}_i$ of a Gaussian distribution, respectively, denoted as $\bm{\mu}_i = f_{\mu}(\bm{z}_i)$ and $\bm{\sigma}_i = \log f_{\sigma}(\bm{z}_i)$. Finally, we sample $\bm{\hat{z}}_i$ from the posterior topic distribution $q(\bm{\hat{z}}_i|\bm{h}_i)$ by using a reparameterization trick as described in~\citet{KingmaW13}, i.e., $\bm{\hat{z_i}} = \bm{\mu}_i + \epsilon\bm{\sigma}_i$. Here, $\epsilon$ is sampled from $\mathcal{N}\left(\mathbf{0},\mathbf{I}\right)$.

Traditional vae-based neural topic models directly feed latent topic representation $\bm{\hat{z}}_i$ into generative network to reconstruct context-aware representation $\bm{h}_i$, which may suffer from the problem of diversity deficiency~\cite{GaoBLLZS19} in capturing multimodal topic information. Therefore, we design a topic-enriched diffusion block (TDB) integrated into the inference network, consisting of topic-enriched diffusion process and topic-enriched denoising process as illustrated in Figure \ref{fig:model}. In the following, we will formulate the TDB integrated into the inference network of TopicDiff in detail.

\textbf{$\bullet$ Topic-enriched Diffusion Process} is leveraged to perturb the latent topic representation $\bm{\hat{z}}_i$ with an infinite number of noise scales. Indexed by a continuous time variable $t \in \left[0, T\right]$, $\bm{\hat{z}}_i \sim p(\bm{\hat{z}}_i)$ and $\bm{\hat{z}}_{i}^{'} \sim p(\bm{\hat{z}}_{i}^{'})$, where $p(\bm{\hat{z}}_i)$ and $p(\bm{\hat{z}}_{i}^{'})$ represent the latent topic distribution and prior distribution in TDB, respectively. Therefore, we have a tractable form to generate representations efficiently, and the topic-enriched diffusion process can be modeled as the solution to a stochastic differential equation (SDE), denoted as:
\begin{equation}
\label{eq:diffusion}
    \mathrm{d}\bm{\hat{z}}_i = f(\bm{\hat{z}}_i, t)\mathrm{d}t + g(t)\mathrm{d}\bm{w}
\end{equation}
where $\bm{w}$ is the standard Wiener process\footnote{https://en.wikipedia.org/wiki/Wiener\_process}. $f(\bm{\hat{z}}_i, t)$ and $g(t)$ represent the vector-valued and scalar function called the drift and diffusion coefficient of $\bm{\hat{z}}_i$, respectively. The SDE has a unique strong solution as long as the coefficients are globally Lipschitz\footnote{https://en.wikipedia.org/wiki/Lipschitz\_continuity} both in state and time.

\textbf{$\bullet$ Topic-enriched Denoising Process} is leveraged to synthesize the latent topic representations $\bm{\hat{z}}_i$ from the prior distribution $p(\bm{\hat{z}}_{i}^{'})$, which also can be seen as a diffusion process, running backwards in time and given by the reverse-time SDE, formulated as:
\begin{equation}
\label{eq:denoising}
    \mathrm{d}{\bm{\hat{z}}_i} = [f(\bm{\hat{z}}_i, t) - g(t)^2 \nabla_{\bm{\hat{z}}_i}\log p(\bm{\hat{z}}_i)]\mathrm{d}t + g(t)\mathrm{d}{\bm{\hat{w}}}
\end{equation}
where $\bm{\hat{w}}$ is the standard Wiener process when time flows backwards from $T$ to $0$, and $\mathrm{d}t$ is an infinitesimal negative time step. Once the score of each marginal distribution, $\nabla_{\bm{\hat{z}}_i}\log p(\bm{\hat{z}}_i)$, is known for all $t$, we can derive the denoising process from Eq.(\ref{eq:denoising}) and simulate it to sample from $p(\bm{\hat{z}}_i)$.

To estimate $\nabla_{\bm{\hat{z}}_i}\log p(\bm{\hat{z}}_i)$, we can train a time-dependent score-based model $s_{\theta}(\bm{\hat{z}}_i, t)$ via a continuous generalization, formulated as: $\theta^* = \arg \min \mathbb{E}_{\bm{\hat{z}}_i} \mathbb{E}_{\bm{\hat{z}}_{i}^{'}|\bm{\hat{z}}_i} \left \| s_{\theta} (\bm{\hat{z}}_i, t) - \nabla_{\bm{\hat{z}}_i} \log p(\bm{\hat{z}}_i) (\bm{\hat{z}}_{i}^{'}|\bm{\hat{z}}_i)\right \|_{2}^{2}$
where $t$ is uniformly sampled over $[0, T]$. With sufficient data and model capacity, score matching ensures that the optimal solution for this formulation, denoted as $s_{\theta^*}(\bm{\hat{z}}_i, t)$, which equals $\nabla_{\bm{\hat{z}}_i}\log p(\bm{\hat{z}}_i)$ for almost all $\bm{\hat{z}}_i$ and $t$. 

\textbf{Generative Network} is leveraged to reconstruct the context-aware representation $\bm{h}_i$ from the latent topic representation $\bm{\hat{z}}_i$ at each time step. Similar to inference network, we employ an MLP decoder to take the sampled latent topic representation $\bm{\hat{z}}_i$ as input and generate reconstructed $\bm{h}_{i}^{'}$ for the context-aware representation $\bm{h}_i$. Besides, generative network also uses Gaussian distributions for both generative prior and variational distribution, while it defines an independent diagonal Gaussian distribution over the embedding space of different modalities. Therefore, generative network simultaneously learns to approximate the conditional probability distribution $p(\bm{h}_{i}^{'}|\bm{\hat{z}}_i)$, which represents the likelihood of generating the reconstructed context-aware representation $\bm{h}_{i}^{'}$ given the latent topic representation $\bm{\hat{z}}_i$.

\subsection{Topic-enriched Learning}
After obtaining the latent topic representation $\bm{\hat{z}}_i$ of each modality, we compute the output representation $\bm{f}_i$ via concatenating $\bm{\hat{z}}_i$ and context-aware representation $\bm{h}_i$ as illustrated in Figure \ref{fig:model}, i.e, $\bm{f}_i=\bm{h}_i \oplus \bm{\hat{z}}_i$, which is used for final emotion detection following basic MCE approaches. 

To train three TopicDiff modules, we minimize the variational upper bound on negative data log-likelihood, consisting of two components: reconstruction loss and Kullback-Leibler (KL) loss. The reconstruction loss is used to minimize the reconstruction error between the context-aware representation $\bm{h}_i$ and the reconstructed $\bm{h}_{i}^{'}$, thereby improving the quality of the reconstructed context-aware representation $\bm{h}_{i}^{'}$, formulated as follows:
\begin{equation}
\mathcal{L}_{rec} = \mathbb{E}_{q(\bm{\hat{z}}_i|\bm{h}_i)} \left[\log p(\bm{h}_{i}^{'}|\bm{\hat{z}}_i)\right]
\end{equation}

And the KL loss is used to minimize the degree of difference between the posterior topic distribution $q(\bm{\hat{z}}_i|\bm{h}_i)$ and the latent topic distribution target distribution $p(\bm{\hat{z}}_i)$, which can be formulated as follows: 
\begin{equation}
\begin{split}
\mathcal{L}_{kl} &= kl(q(\bm{\hat{z}}_i|\bm{h}_i)||p(\bm{\hat{z}}_i))
\end{split}
\end{equation}

Therefore, we jointly train the basic MCE approaches alongside three TopicDiff modules, and the total training objective can be denoted as:
\begin{equation}
    \mathcal{L}_{total} \! \!= \! \! \mathcal{L}_{mce} \! \! + \! \!  \alpha\sum\nolimits_{(\boldsymbol{a},\boldsymbol{v},\boldsymbol{l})}\mathcal{L}_{rec} \! \! + \! \! \beta\sum\nolimits_{(\boldsymbol{a},\boldsymbol{v},\boldsymbol{l})}\mathcal{L}_{kl}
\end{equation}
where $\sum_{(\boldsymbol{a},\boldsymbol{v},\boldsymbol{l})}$ represents the sum of losses from corresponding three modalities. Besides, $\alpha$ and $\beta$ are hyper-parameters of weight to balance the losses between basic MCE approaches and three TopicDiff modules.

\setlength{\tabcolsep}{1.8pt}
\begin{table}[t]
\renewcommand{\arraystretch}{1.1}
\addtolength{\tabcolsep}{1pt}
\begin{center}
\scalebox{0.75}{
\begin{tabular}{c|ccc|ccc}
\toprule[1.2pt]
\multirow{2}{*}{Dataset} & \multicolumn{3}{c|}{Conversations} & \multicolumn{3}{c}{Utterances}  \\ \cline{2-7} 
                         & Train+Val   & Test & Total & Train+Val  & Test  & Total  \\ \hline
M3ED$^*$                     & 809   & 181  & 990   & 19,702  & 4,747 & 24,449 \\
MELD                     & 1,153     & 280  & 1,433 & 11,098   & 2,610 & 13,708 \\
IEMOCAP                     & 120    & 31  & 151 & 5,810 & 1,623 & 7,433 \\
\bottomrule[1.2pt]
\end{tabular}
}
\caption{The statistics of our constructed topic-density M3ED$^*$ and two public topic-sparsity MELD, IEMOCAP datasets. $*$ denotes this dataset is different from the original.}
\label{tab:dataset} 
\end{center}
\end{table}

\setlength{\tabcolsep}{1.8pt}
\begin{table*}[t]
\renewcommand{\arraystretch}{1.2}
\addtolength{\tabcolsep}{3.8pt}
\begin{center}
\scalebox{0.75}{
    \begin{tabular}{c|cccccccc|c|c}
\toprule[1.2pt]
                           & \multicolumn{8}{c|}{M3ED$^*$}                                                                                      & MELD & IEMOCAP  \\ \cline{2-11} 
\multirow{-2}{*}{Approach} & Happy & Neutral & Sad   & Disgust & Angry & Fear  & \multicolumn{1}{c|}{Surprise}                      & W-F1  & W-F1    & W-F1  \\ \hline
DialogueCRN                & 54.38 & 67.75   & 54.27 & 34.59   & 70.74 & 12.10 & \multicolumn{1}{c|}{55.55}                         & 61.32 & 54.32$^\dagger$   & 65.04$^\sharp$ \\
\rowcolor[HTML]{EFEFEF} 
+ TopicDiff                     & 56.22($\uparrow$) & 72.21($\uparrow$)   & 55.06($\uparrow$) & 38.92($\uparrow$)   & 73.41($\uparrow$) & 28.45($\uparrow$) & \multicolumn{1}{c|}{\cellcolor[HTML]{EFEFEF}55.08($\downarrow$)} & 64.49($\uparrow$) & 55.36($\uparrow$)   & 66.05($\uparrow$) \\
+ TopicDiff w/o TDB             & 54.13 & 68.60   & 54.45 & 37.94   & 72.24 & 27.83 & \multicolumn{1}{c|}{53.21}                         & 62.36 & 54.43       & 65.31     \\ \hline
MMGCN                      & 58.83 & 69.00   & 56.68 & 34.31   & 69.61 & 23.47 & \multicolumn{1}{c|}{54.17}                         & 62.51 & 57.26$^\dagger$   & 66.22$^\sharp$ \\
\rowcolor[HTML]{EFEFEF} 
+ TopicDiff                     & 62.70($\uparrow$) & 73.03($\uparrow$)   & 57.80($\uparrow$) & 38.98($\uparrow$)   & 72.08($\uparrow$) & 33.02($\uparrow$) & \multicolumn{1}{c|}{\cellcolor[HTML]{EFEFEF}55.95($\uparrow$)} & 65.72($\uparrow$) & 58.26($\uparrow$)   & 67.02($\uparrow$) \\
+ TopicDiff w/o TDB             & 60.52 & 72.10   & 58.11 & 36.55   & 71.39 & 0.8   & \multicolumn{1}{c|}{43.46}                         & 63.94 & 57.63       & 66.47     \\ \hline
COGMEN                    & 59.25 & 71.20   & 56.98 & 40.20   & 73.50 & 22.94 & \multicolumn{1}{c|}{58.93}                         & 64.88 & 52.29$^\dagger$   & 64.56$^\dagger$ \\
\rowcolor[HTML]{EFEFEF} 
+ TopicDiff                      & 60.95($\uparrow$) & 72.84($\uparrow$)   & 60.180($\uparrow$) & 38.18($\downarrow$)   & 74.32($\uparrow$) & 25.63($\uparrow$) & \multicolumn{1}{c|}{\cellcolor[HTML]{EFEFEF}60.86($\uparrow$)} & 66.39($\uparrow$) & 53.54($\uparrow$)   & 65.48($\uparrow$) \\
+ TopicDiff w/o TDB             &59.45 & 71.64   & 57.29 & 39.83   & 73.98 & 20.37 & \multicolumn{1}{c|}{61.56}                         & 65.26 & 52.76       & 64.91     \\ \hline
MM-DFN                    & 62.29 & 76.81   & 60.72 & 43.58   & 74.99 & 14.77 & \multicolumn{1}{c|}{61.88}                         & 68.58 & 57.54$^\dagger$   & 65.66$^\dagger$ \\
\rowcolor[HTML]{EFEFEF} 
+ TopicDiff                      & 63.69($\uparrow$) & 77.78($\uparrow$)   & 61.60($\uparrow$) & 45.66($\uparrow$)   & 76.47($\uparrow$) & 38.02($\uparrow$) & \multicolumn{1}{c|}{\cellcolor[HTML]{EFEFEF}62.140($\uparrow$)} & 70.06($\uparrow$) & 58.42($\uparrow$)   & 66.52($\uparrow$) \\
+ TopicDiff w/o TDB             & 62.78 & 77.57   & 59.903 & 44.41   & 75.76 & 24.52 & \multicolumn{1}{c|}{60.55}                         & 69.10 & 57.97       & 65.85     \\ \hline
GCNet                   & 46.65 & 72.24   & 47.09 & 27.40   & 66.77 & 3.73 & \multicolumn{1}{c|}{38.40}                         & 59.02 & -   & 56.18$^\sharp$ \\
\rowcolor[HTML]{EFEFEF} 
+ TopicDiff                      & 51.54($\uparrow$) & 71.09($\downarrow$)   & 51.21($\uparrow$) & 36.46($\uparrow$)   & 71.42($\uparrow$) & 8.92($\uparrow$) & \multicolumn{1}{c|}{\cellcolor[HTML]{EFEFEF}45.63($\uparrow$)} & 61.71($\uparrow$) & -   & 57.80($\uparrow$) \\
+ TopicDiff w/o TDB             & 50.04 & 70.97   & 49.64 & 24.53   & 69.39 & 4.68 & \multicolumn{1}{c|}{41.52}                         & 59.78 & -       & 56.78      \\ 
\bottomrule[1.2pt]
\end{tabular}
}
\setlength{\abovecaptionskip}{0.5 ex}
  \setlength{\belowcaptionskip}{-2 ex}
\caption{Comparison of several MCE approaches with and without our TopicDiff approach on our constructed topic-density M3ED$^*$ and two public topic-sparsity MELD, IEMOCAP datasets. The grey rows highlights the results of TopicDiff applied to MCE approaches, where $\uparrow$ and $\downarrow$ represent improvement and reduction respectively, and $\uparrow$ indicates better performance for all metrics. Since TopicDiff is model-agnostic, for a fair comparison, we reproduce all the baselines under the same experimental environment and independently run five times with random seeds, taking their average as final results to report. Particularly, these reproduced results with symbol $\sharp$ are close to those reported in \citet{HuLZJ20} and \citet{LianCSLT23}, while those reproduced results with $\dagger$ are slightly lower than results reported in their original papers due to different experimental environments. The symbol - denotes the results are not available, because GCNet is not suitable to multiple speakers in MELD dataset as reported in \cite{LianCSLT23}.}
\label{tab:result}
\end{center}
\end{table*}

\section{Experimental Settings}  
\subsection{Datasets and Baselines}
For datasets, we empirically evaluate our TopicDiff approach on our constructed topic-density M3ED$^*$ and two public topic-sparsity MELD, IEMOCAP datasets. \textbf{MELD}~\cite{PoriaHMNCM19} is a topic-sparsity dataset containing multi-speaker conversations from the only one popular TV series \emph{Friends}. \textbf{IEMOCAP}~\cite{BussoBLKMKCLN08} is also a topic-sparsity dataset containing dyadic conversations where actors perform improvisations or scripted scenarios.  \textbf{M3ED$^*$} is a topic-density dataset containing dyadic conversations from 56 different TV series, thus surpassing MELD and IEMOCAP in topic richness. We construct this topic-density M3ED$^*$ dataset based on the original M3ED dataset\footnote{https://github.com/AIM3-RUC/RUCM3ED} to evaluate the effectiveness of multimodal topic information. Specifically, for the construction of M3ED$^*$, we partition the TV series into 56, 10, and 20 for the training, validation and test sets respectively, and maintain the 7:1:2 ratio of utterances, which is different from the original M3ED dataset in \citet{ZhaoZ0LJW022}. To focus on evaluating the effectiveness of TopicDiff towards mining multi-topic information, we ensure that the TV domains in test set are seen in the training set, which furthest excludes the influence of adaptation problem stemming from different TV domains. All the three datasets include aligned acoustic, vision and language three modalities, and a comprehensive statistical summary is reported in Table \ref{tab:dataset}.

For baselines, we choose DialogueCRN~\cite{HuWH20}, MMGCN~\cite{HuLZJ20}, COGMEN~\cite{JoshiBJSM22}, MM-DFN~\cite{HuHWJM22} and GCNet~\cite{LianCSLT23} five MCE approaches to evaluate the effectiveness of our TopicDiff approach. We aim to compare the performance differences of these MCE approaches with and without TopicDiff. It is worth noting that these approaches are all multimodal approaches, rather than only using language modality. Since existing single-text conversation approaches, such as COSMIC~\cite{GhosalMGMP20} and MuCDN~\cite{ZhaoZ022}, finetune pre-trained models like RoBERTa~\cite{abs-1907-11692} and incorporate external knowledge (e.g., commonsense and events), we consider it is unfair to compare with single-text conversation approaches. Because the above approaches have different experimental settings (e.g., different features or partitions in datasets, different evaluation metrics, different devices etc.), for a fair and thorough comparison, we reproduce these approaches in our experiments. 

\subsection{Implementation Details and Metrics}
Since our TopicDiff approach is model-agnostic, for a fair comparison, we reproduce all the baselines on our topic-density M3ED$^*$ and two public topic-sparsity MELD, IEMOCAP datasets under the same experimental environment (i.e., all the approaches are reproduced with PyTorch on a machine with NVIDIA GeForce RTX 3090, Intel(R) Xeon(R) E5-2650 v4 CPU (2.20 GHz), CUDA version 11.7, and PyTorch 1.7.1 library with python 3.6.13 on Ubuntu 20.04.1 LTS.). Besides, the hyper-parameters of these baselines reported by their public papers are still adopting the same setting, and the others are tuned according to the validation set. Furthermore, we conduct five independent runs with random seeds for each baseline with and without TopicDiff, taking their average as the final results to report. Particularly, the reproduced results of DialogueCRN, MMGCN and GCNet on IEMOCAP could achieve close results as reported in their corresponding papers. However, the reproduced results of other baselines on two public topic-sparsity MELD, IEMOCAP datasets are slightly lower than those results reported in their corresponding papers. For our TopicDiff approach in experimental settings, we use a simple VAE consisting of MLP with linear layers as inference and generative networks, and leverage NCSN~\cite{SongE19} as score-based diffusion backbone. Specifically, we adopt Adam as the optimizer with two initial learning rates (1e-4, 1e-5) and L2 weight decays (1e-4, 1e-4) for VAE and NCSN to update training parameters, respectively. The dropout rate is set to 0.25 for VAE. The dimension of hidden state after VAE is set to 20, which also represents the number of topics. The implementation details of NCSN are referenced in their publicly available code\footnote{https://github.com/ermongroup/ncsn}. In addition, we set the weights $\alpha$ and $\beta$ to (0.5, 0.5) to balance the losses between MCE and TopicDiff. All the approaches are trained for a maximum of 200 epochs, and stopped if the validation loss does not decrease for 20 consecutive epochs. To facilitate the corresponding research in this direction, all codes together with datasets are released at Github\footnote{Github link will come soon.}. 

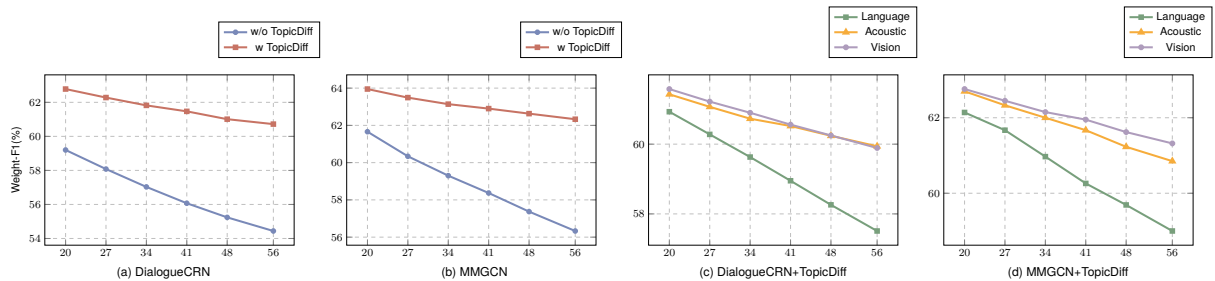
\begin{figure*}[t]
\centering
\definecolor{color1}{RGB}{121,137,184}
\definecolor{color2}{RGB}{199,115,100}
\definecolor{color3}{RGB}{120,159,124}
\definecolor{color4}{RGB}{246,172,59}
\definecolor{color5}{RGB}{174,156,188}
\resizebox{\linewidth}{!}{
\subfloat[DialogueCRN]{
\begin{tikzpicture}
    \scalefont{0.8}
    \begin{axis}[
    sharp plot,
    xmode=normal,
    ylabel=Weight-F1(\%),
    width=8cm,
    height=6cm,
    xtick={20,27,34,41,48,56},
    ytick={52,54,56,58,60,62},
    ylabel near ticks,
    xmajorgrids=true,
    ymajorgrids=true,
    grid style=dashed,
    legend style={at={(0.9,1.1)},anchor=south}
    ]
      \addplot+[semithick,mark options={scale=0.65},color=color1, line width=1.5pt] plot coordinates{
      (20,59.20)
      (27,58.08)
      (34,57.03)
      (41,56.07)
      (48,55.24)
      (56,54.44)
      };
      \addlegendentry{w/o TopicDiff}
      \addplot+[semithick,mark options={scale=0.65},color=color2, line width=1.5pt] plot coordinates{
      (20,62.78)
      (27,62.28)
      (34,61.82)
      (41,61.47)
      (48,61.01)
      (56,60.72)
      };
      \addlegendentry{w TopicDiff}
    \end{axis}
\end{tikzpicture}
}
\hfill
\subfloat[MMGCN]{
\begin{tikzpicture}
    \scalefont{0.8}
    \begin{axis}[
    sharp plot,
    xmode=normal,
    width=8cm,
    height=6cm,
    xtick={20,27,34,41,48,56},
    ytick={54,56,58,60,62,64},
    xmajorgrids=true,
    ymajorgrids=true,
    grid style=dashed,
    legend style={at={(0.9,1.1)},anchor=south}
    ]
      \addplot+[semithick,mark options={scale=0.65},color=color1, line width=1.5pt] plot coordinates{
      (20,61.66)
      (27,60.34)
      (34,59.30)
      (41,58.37)
      (48,57.37)
      (56,56.33)
      };
      \addlegendentry{w/o TopicDiff}
      \addplot+[semithick,mark options={scale=0.65},color=color2, line width=1.5pt] plot coordinates{
      (20,63.95)
      (27,63.49)
      (34,63.14)
      (41,62.90)
      (48,62.63)
      (56,62.33)
      };
      \addlegendentry{w TopicDiff}
    \end{axis}
\end{tikzpicture}
}
\subfloat[DialogueCRN+TopicDiff]{
\begin{tikzpicture}
    \scalefont{0.8}
    \begin{axis}[
    sharp plot,
    xmode=normal,
    width=8cm,
    height=6cm,
    xtick={20,27,34,41,48,56},
    ytick={52,54,56,58,60,62,64},
    xmajorgrids=true,
    ymajorgrids=true,
    grid style=dashed,
    legend style={at={(0.9,1.1)},anchor=south}
    ]
      \addplot+[semithick,mark=square*,mark options={scale=0.65},color=color3, line width=1.5pt] plot coordinates{
      (20,60.93)
      (27,60.28)
      (34,59.63)
      (41,58.95)
      (48,58.26)
      (56,57.51)
      };
      \addlegendentry{Language}
      \addplot+[semithick,mark=triangle*,mark options={scale=1},color=color4, line width=1.5pt] plot coordinates{
      (20,61.43)
      (27,61.07)
      (34,60.73)
      (41,60.52)
      (48,60.24)
      (56,59.94)
      };
      \addlegendentry{Acoustic}
      \addplot+[semithick,mark options={scale=0.7},color=color5, line width=1.5pt] plot coordinates{
      (20,61.58)
      (27,61.22)
      (34,60.90)
      (41,60.56)
      (48,60.25)
      (56,59.89)
      };
      \addlegendentry{Vision}
    \end{axis}
\end{tikzpicture}
}
\hfill
\subfloat[MMGCN+TopicDiff]{
\begin{tikzpicture}
    \scalefont{0.8}
    \begin{axis}[
    sharp plot,
    xmode=normal,
    width=8cm,
    height=6cm,
    xtick={20,27,34,41,48,56},
    ytick={54,56,58,60,62,64},
    xmajorgrids=true,
    ymajorgrids=true,
    grid style=dashed,
    legend style={at={(0.9,1.1)},anchor=south}
    ]
      \addplot+[semithick,mark=square*,mark options={scale=0.65},color=color3, line width=1.5pt] plot coordinates{
      (20,62.14)
      (27,61.67)
      (34,60.97)
      (41,60.26)
      (48,59.69)
      (56,59.00)
      };
      \addlegendentry{Language}
      \addplot+[semithick,mark=triangle*,mark options={scale=1},color=color4, line width=1.5pt] plot coordinates{
      (20,62.70)
      (27,62.33)
      (34,62.00)
      (41,61.67)
      (48,61.23)
      (56,60.85)
      };
      \addlegendentry{Acoustic}
      \addplot+[semithick,mark options={scale=0.7},color=color5, line width=1.5pt] plot coordinates{
      (20,62.76)
      (27,62.45)
      (34,62.15)
      (41,61.95)
      (48,61.62)
      (56,61.32)
      };
      \addlegendentry{Vision}
    \end{axis}
\end{tikzpicture}
}
}
\setlength{\abovecaptionskip}{0.5 ex}
  \setlength{\belowcaptionskip}{-2 ex}
\caption{Four line charts to study the robustness of our TopicDiff approach and different modal topic information with the change of topic-density degree via maintaining the total number of training set and varying the numbers of TV series, where the x-axis represents the numbers of TV series. Line charts of (a) and (b) show the performance trend on two MCE approaches (with/without TopicDiff). And line charts of (c) and (d) illustrate the performance trend on acoustic, vision and language topic information via two MCE approaches with TopicDiff. All the line charts are conducted on our constructed topic-density M3ED$^*$ dataset and evaluated on W-F1 metric.}
\label{fig:analysis}
\end{figure*}

Following recent works~\cite{HuWH20}, the performance is evaluated via \emph{Weighted-Average F1} (W-F1) on our constructed topic-density M3ED$^*$ and two public topic-sparsity MELD, IEMOCAP datasets. Besides, we report the F1-score of each emotion on our topic-density M3ED$^*$ dataset. Moreover, $t$-test\footnote{https://docs.scipy.org/doc/scipy/reference/stats.html} is used to evaluate the significance of performance differences.

\section{Results and Discussions}  
\subsection{Experimental Results}
Table \ref{tab:result} illustrates the comparative results for MCE. From the table, we can see that our TopicDiff approach applied to various MCE approaches significantly outperforms those without TopicDiff on our constructed topic-density M3ED$^*$ and two public topic-sparsity MELD, IEMOCAP datasets. For instance, TopicDiff yields an improvement of 3.21\% in terms of W-F1 over the MMGCN on our constructed topic-density M3ED$^*$ dataset (p-value $<$ 0.01), and 1\% and 1.2\% on two public topic-sparsity MELD, IEMOCAP datasets respectively (p-value $<$ 0.05). This indicates that our TopicDiff approach is more effective on topic-density scenarios, and further encourages us to consider incorporating multimodal topic information in MCE.

\setlength{\tabcolsep}{1.8pt}
\begin{table}[htb]
\renewcommand{\arraystretch}{1.1}
\addtolength{\tabcolsep}{6pt}
\begin{center}
\resizebox{\linewidth}{!}{%
\begin{tabular}{@{\hspace{8pt}}c@{\hspace{8pt}}c@{\hspace{8pt}}c|cc|ll}
\toprule[1.2pt] 
Language & Acoustic & Vision &  \multicolumn{2}{c|}{DialogueCRN}          & \multicolumn{2}{c}{MMGCN}                 \\ \hline
\checkmark &       &        & 62.34 & \cellcolor[HTML]{EFEFEF}\textbf{+1.02} & 63.69 & \cellcolor[HTML]{EFEFEF}\textbf{+1.18} \\
     & \checkmark     &        & 62.78 & \cellcolor[HTML]{EFEFEF}\textbf{+1.46} & 64.03 & \cellcolor[HTML]{EFEFEF}\textbf{+1.52} \\
     &       & \checkmark      & 62.82 & \cellcolor[HTML]{EFEFEF}\textbf{+1.50} & 64.09 & \cellcolor[HTML]{EFEFEF}\textbf{+1.58} \\
\checkmark    & \checkmark     &       & 63.13 & \cellcolor[HTML]{EFEFEF}\textbf{+1.81} & 64.43 & \cellcolor[HTML]{EFEFEF}\textbf{+1.92} \\
\checkmark    &       & \checkmark     & 63.18 & \cellcolor[HTML]{EFEFEF}\textbf{+1.86} & 64.42 & \cellcolor[HTML]{EFEFEF}\textbf{+1.91} \\
     & \checkmark     & \checkmark     & 63.55 & \cellcolor[HTML]{EFEFEF}\textbf{+2.23} & 64.76 & \cellcolor[HTML]{EFEFEF}\textbf{+2.25} \\
\checkmark    & \checkmark     & \checkmark     & 64.49 & \cellcolor[HTML]{EFEFEF}\textbf{+3.17} & 65.72 & \cellcolor[HTML]{EFEFEF}\textbf{+3.21} \\ 
\bottomrule[1.2pt]
\end{tabular}
}
\setlength{\abovecaptionskip}{0.5 ex}
  \setlength{\belowcaptionskip}{-2 ex}
\caption{The effectiveness study of various modal topic information in MCE, where \checkmark means that we capture the current modal topic information. The grey columns show the difference of MCE approaches with and without various modal topic information via TopicDiff. All the experiments are conducted on our constructed topic-density M3ED$^*$ dataset and evaluated on W-F1 metric.}
\label{tab:modality}
\end{center}
\end{table}

Furthermore, we report the results for individual emotions on our constructed topic-density M3ED$^*$ dataset, where TopicDiff boosts the performance of most individual emotions. Impressively, TopicDiff applied to MMGCN improves the performance of all individual emotions, with a significant improvement up to 9.55\% for the \emph{fear} emotion (p-value $<$ 0.01). This highlights the capability of TopicDiff to boost the detection of challenging emotions.

\subsection{Ablation Study of TDB}
In order to demonstrate the effectiveness of integrating the diffusion model into neural topic model for fully capturing multimodal topic information, we conduct ablation studies on our constructed topic-density M3ED$^*$ and two public topic-sparsity MELD, IEMOCAP datasets. Specifically, we remove the topic-enriched diffusion block (TDB) from our TopicDiff approach, and apply the remaining components (i.e., neural topic model) to various MCE approaches. We present the results (i.e., + TopicDiff w/o TDB) in Table \ref{tab:result} for a clear and intuitive comparison with our primary results. The results show that while the removal of TDB could improve performance, it is not as significant as that achieved with TopicDiff. This indicates that integrating the diffusion model into neural topic model could mitigate diversity deficiency problem, thereby fully capturing multimodal topic information in MCE.

\begin{figure*}[t]
\begin{center}
    \subfloat{
 \includegraphics[scale=0.55]{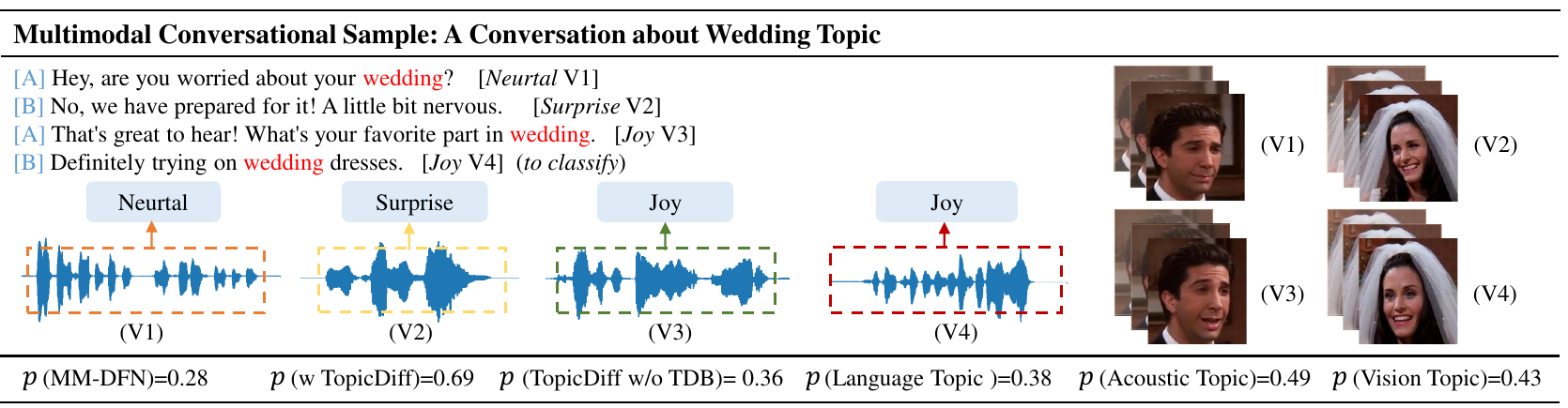}}
\caption{A multimodal conversational sample includes utterances comprising acoustic spectrums, video frames, language, and ground-truth emotions, alongside the probabilities of ground-truth emotion \emph{joy} on V4 predicted by various approaches. Language/Acoustic/Vision Topic denotes the utilization of TopicDiff to capture the corresponding modal topic information.}
\label{fig:case}
\end{center}
\end{figure*}

\subsection{Effectiveness Study of Modal Topics}
Table \ref{tab:modality} presents a comprehensive evaluation of TopicDiff in exploring the contributions of various modal topic information on our constructed topic-density M3ED$^*$ dataset. Specifically, we randomly choose DialogueCRN and MMGCN with TopicDiff to capture various combinations of modal topic information, and calculate the performance improvements between baselines with and without the incorporation of various modal topic information. From the table, we can see that the incorporation of topic information from various modalities still yields improvements in emotion detection, and the use of three modal topic information outperforms that of one or two modal topic information. This demonstrates the importance of topic information in conversation tasks, and further highlights the potential effectiveness of multimodal topic information.

Moreover, we have observed an interesting experimental phenomenon wherein incorporating acoustic or vision topic information leads to better performance than language. For instance, the incorporation of acoustic and vision topic information yields improvements of 0.44\% and 0.48\% compared to language via using DialogueCRN respectively. This observation demonstrates that compared to language, acoustic and vision topic information is more discriminative, which is reasonable since acoustic and vision signals are more natural and direct than language, making their associated topic information more discriminative to emotion. 

\subsection{Robustness Study of Modal Topics}
To investigate the impact of topic-density degree on the effectiveness of our TopicDiff approach, we conduct experiments that involved sampling different numbers of TV series and analyzing the resulting changes in performance. Specifically, we randomly select 5,000 utterances from the training set, while varying the number of TV series to 20, 27, 34, 41, 48, and 56. Subsequently, we randomly choose DialogueCRN and MMGCN, with and without TopicDiff, to analyze the performance trend concerning changes in the topic-density degree. As illustrated in Figure \ref{fig:analysis} (a) and (b), all approaches show a decrease in performance due to the reduction in the number of samples from the same TV domain, while maintaining the total number (i.e., 5,000) of the training set\footnote{To furthest exclude the influence of domain adaptation, similar to the construction of M3ED$^*$, we ensure that TV domains in test set are seen in the training set.} and increasing topic-density, necessitating the use of multimodal topic information to sustain performance. We observe a significant decline in performance for the two MCE approaches, while TopicDiff exhibits a slight and stable decline. This demonstrates that our TopicDiff approach is robust since it incorporates additional multimodal topic information.

To further explore the performance difference of various modal topic information under the change of topic-density degree, we randomly choose DialogueCRN and MMGCN with TopicDiff to capture acoustic, vision or language topic information. We plot the trend of performance changes in Figure \ref{fig:analysis} (c) and (d). The results reveal that TopicDiff, with either acoustic or vision topic information, maintains a more stable performance compared to language, even with changes in topic-density degree. This interesting observation indicates that acoustic and vision topic information is more robust than language, which is reasonable since the emotional clues expressed by acoustic and vision tend to be more consistent across different topics than language. For example, the facial expressions of \emph{smile} are always similar in different topics, thus making them have better robustness.

\subsection{Qualitative Study of TopicDiff}
As shown in Figure \ref{fig:case}, we randomly choose DialogueCRN with and without TopicDiff and leverage TopicDiff to capture the topic information of language, acoustic and vision to predict the probability of ground-truth emotion \emph{joy} on V4. From this figure, we can see that: 1) Predicting the emotion of V4 in the absence of context is challenging. Moreover, the use of global language topic clues (i.e., ``\emph{worried about}'', ``\emph{nervous}'') may lead to negative predictions. Thus, it is essential to combine acoustic and vision topic clues (i.e., \emph{joy tones}, \emph{smile face}) for predicting the emotion of V4 precisely. This highlights the importance of multimodal topic information in conversations. 2)  DialogueCRN with TopicDiff provides a higher probability of \emph{joy} emotion on V4 compared to DialogueCRN, indicating the effectiveness of TopicDiff in capturing multimodal topic information. 3) TopicDiff without TDB provides a lower probability compared to TopicDiff, indicating the effectiveness of integrating the diffusion model to capture multimodal topic information. 4) The prediction probabilities of acoustic or vision topic are obviously higher than language topic, indicating that acoustic and vision topic information is more discriminative compared to language.
\section{Conclusion}
In this paper, we propose a model-agnostic \textbf{Topic}-enriched \textbf{Diff}usion (TopicDiff) approach to MCE, which integrates the diffusion model into neural topic model for mitigating the problem of semantic diversity deficiency in capturing multimodal topic information. Detailed experiments on our topic-density M3ED$^*$ and two public topic-sparsity MELD, IEMOCAP datasets demonstrate the effectiveness of our TopicDiff approach in MCE. Furthermore, we observe an interesting finding that acoustic and vision topic information is more discriminative and robust compared to language. In our future work, we would like to improve the emotion detection by incorporating more information, such as speaker personality and posture information. Besides, we would like to leverage large-scale multimodal video pre-training to enhance the representation ability of basic modalities. Moreover, we intend to transfer TopicDiff to other conversation based multimodal tasks, such as multimodal psychological counseling and conversational depression detection. 
\section*{Ethics Statement}
\textbf{Data Disclaimer.}
We construct a topic-density M3ED$^*$ dataset based on M3ED, which is available at \url{https://github.com/AIM3-RUC/RUCM3ED}. M3ED dataset is widely used by other academics and are typically accessible to the public. Therefore, our proposed dataset does not involve any sensitive information that may harm others.

\noindent\textbf{Human Construction.} 
When recruiting constructors for this study, we claim that all potential constructors are free to choose whether they want to participate, and they can withdraw from the study anytime without any negative repercussions. Thus, the establishment of our topic-density M3ED$^*$ dataset is compliant with ethics.

\section*{Acknowledgments}
We thank our anonymous reviewers for their helpful comments. This work was supported by three NSFC grants, i.e., No.62006166, No.62376178 and No.62076175. This work was also supported by a Project Funded by the Priority Academic Program Development of Jiangsu Higher Education Institutions (PAPD).

\section*{References}
\bibliographystyle{lrec-coling2024-natbib}
\bibliography{lrec-coling2024-example}

\bibliographystylelanguageresource{lrec-coling2024-natbib}
\bibliographylanguageresource{languageresource}

% \clearpage
% \input{contents/appendix}

\end{document}